\documentclass[conference]{IEEEtran}
\IEEEoverridecommandlockouts
\usepackage{cite}
\usepackage{amsmath,amssymb,amsfonts,bm,amsthm,balance}
\usepackage{algorithmic,algorithm}
\usepackage{graphicx}
\usepackage{subcaption}
\usepackage{textcomp}
\usepackage{xcolor,color,colortbl}
\def\BibTeX{{\rm B\kern-.05em{\sc i\kern-.025em b}\kern-.08em
    T\kern-.1667em\lower.7ex\hbox{E}\kern-.125emX}}

 \DeclareFontFamily{OT1}{pzc}{}
\DeclareFontShape{OT1}{pzc}{m}{it}{<-> s * [1.10] pzcmi7t}{}
\DeclareMathAlphabet{\mathpzc}{OT1}{pzc}{m}{it}

\newcommand{\grad}[1]{\nabla_{#1^{\mathsf{T}}} }
\newcommand{\w}{\bm{w}}
\newcommand{\si}{\bm{s}_i}

\newcommand{\eqdef}{\:\overset{\Delta}{=}\:}
\DeclareMathOperator*{\argmin}{argmin}
\newcommand{\Li}{\mathcal{\bm{L}}_i}

\definecolor{Gray}{gray}{0.8}
\definecolor{LightCyan}{rgb}{0.88,1,1}

\newtheorem{theorem}{Theorem}
\newtheorem{assumption}{Assumption}
\newtheorem{lemma}{Lemma}

\begin{document}

\title{Dynamic Federated Learning}


\author{\IEEEauthorblockN{Elsa Rizk \IEEEauthorrefmark{1}, Stefan Vlaski \IEEEauthorrefmark{2}, and Ali H. Sayed \IEEEauthorrefmark{3}}
\IEEEauthorblockA{Institute of Electrical Engineering, Ecole Polytechnique Federal de Lausanne\\
Switzerland \\
Email: \IEEEauthorrefmark{1}elsa.rizk@epfl.ch, \IEEEauthorrefmark{2}stefan.vlaski@epfl.ch, \IEEEauthorrefmark{3}ali.sayed@epfl.ch}}

\maketitle

\begin{abstract}
Federated learning has emerged as an umbrella term for centralized coordination strategies in multi-agent environments. While many federated learning architectures process data in an online manner, and are hence adaptive by nature, most performance analyses assume static optimization problems and offer no guarantees in the presence of drifts in the problem solution or data characteristics. We consider a federated learning model where at every iteration, a random subset of available agents perform local updates based on their data. Under a non-stationary random walk model on the true minimizer for the aggregate optimization problem, we establish that the performance of the architecture is determined by three factors, namely, the data variability at each agent, the model variability across all agents, and a tracking term that is inversely proportional to the learning rate of the algorithm. The results clarify the trade-off between convergence and tracking performance.
\end{abstract}


\section{Introduction}
\noindent We consider a collection of $K$ agents dispersed in space. Each agent $k$ has access to $N_k$ data points denoted by $\{x_{k,n}\}$, where the subscript $k$ denotes the agent index and the subscript $n$ denotes the sample index within the dataset. The objective is to seek the minimizer of the aggregate risk:
\begin{equation}\label{eq:globalProb}
w^o \triangleq  \argmin_{w\in\mathbb{R}^{M}} 
\frac{1}{K} \sum_{k=1}^K P_k(w),
\end{equation}
where the individual risks at the local agents are in turn defined as sample averages over their loss values, i.e.,
\begin{equation}
P_k(w)\triangleq \frac{1}{N_k}\sum_{n=1}^{N_k} Q_k(w; x_{k,n}).
\end{equation}
We will allow the sought-after model $w^o$ to drift with time and often write $w_i^o$ instead of just $w^o$ to highlight this possibility. Here, the subscript $i$ refers to a time index. The drift in $w^o$ is often the result of variations in the statistical properties of the data $\{x_{k,n}\}$ at the agents, which can change with time as well. 

Strategies for the pursuit of solutions to~\eqref{eq:globalProb} can generally be divided into one of two classes: distributed architectures with a fusion center that collects all data centrally for processing
\cite{Duchi11,Zinkevich10,Bertsekas96anew}, and fully decentralized strategies that rely on local information exchanges among neighbouring agents over a graph \cite{sayed2014adaptation,chen2012limiting,Nedic09,Duchi12a}. 
Federated learning~\cite{mcmahan16,li2018federated,smith2017federated,caldas2018expanding,stattler19,konen2016federated,
mohri2019agnostic,corinzia2019variational,khodak2019adaptive,chen2018federated,
bonawitz2017practical,geyer2017differentially,mcmahan2017learning} offers a midterm solution where data is collected locally at the agents and some processing is also performed locally, while global information is shared between a central processor and the dispersed agents.  The architecture helps reduce the amount of communication rounds between the central processor and the agents.

\subsection{Related Works}
Several recent works have examined the convergence behavior of federated learning. They, however, vary in the assumptions on the number of participating agents (all or partial), nature of data (IID or not), operation (synchronous or asynchronous), nature of risk function (convex or non-convex), and bounded conditions on gradient noise \cite{jiang2018linear,khaled2019analysis,stich2018local,Wang2018CooperativeSA,Zhou_2018,yu2018parallel,wang2018adaptive,yu19,xie2019asynchronous,
li2019convergence,li2018federated,smith2017federated,liu2019clientedgecloud}  -- see Table 1. 

Some other works examine the convergence behaviour of a modified version of the original FedAvg algorithm; in \cite{li2018federated} FedAvg is modified to include non-uniform epoch sizes among the agents; the authors then provide a convergence proof of the new algorithm, called FedProx, independent of the local solvers. They assume non-convex cost functions with non-IID data, and they quantify the statistical heterogeneity by a dissimilarity measure based on the randomized Kaczmarz method \cite{Kaczmarz, Strohmer2008} for solving linear system of equations. However, this proof fails to encapsulate the convergence of FedAvg. In \cite{liu2019clientedgecloud}, a hierarchical version of FedAvg is developed where model aggregation occurs at several levels, and the convergence under non-IID settings for both convex and non-convex functions is studied. The authors of \cite{smith2017federated} introduce multi-task federated learning to deal with statistical heterogeneity and show their algorithm converges for convex cost functions. They introduce the MOCHA algorithm that extends CoCoA \cite{jaggi2014communication} to deal with the challenges introduced by the federated setting, such as stragglers. 

While related works have considered dynamic optimization algorithms under varying algorithmic frameworks~\cite{Tang,Enyioha,Simonetto,Ling,TOWFIC2013138,wotao19,dall2019optimization,bastianello2020distributed}, we focus in this work on the original FedAvg algorithm with varying step-sizes and under a non-stationary environment.

\section{Algorithm Derivation}

\begin{table*}[htbp]
\vspace*{0.1cm}
\caption{List of references on the convergence analysis of federated learning under different assumptions. This work is the only one to tackle the 3 challenges of federated learning.}
\begin{center}
\begin{tabular}{|c|c|c|c|c|c|c|}
\hline
\rowcolor{Gray}
\textbf{References} & \textbf{Algorithm}& \textbf{Function Type}& \textbf{Data Heterogeneity} &\textbf{Operation} & \textbf{Agent Participation} & \textbf{Other Assumptions} \\
\hline
\cite{khaled2019analysis} & dist. gradient descent & convex &  \textbf{\textit{non-IID}} & synchronous & full & smooth\\
\hline
\cite{stich2018local} & dist. SGD & convex & IID & synchronous & full & smooth \\
\hline
\cite{Wang2018CooperativeSA, Zhou_2018} & dist. SGD & non-convex & IID & synchronous & full & smooth \\
\hline
\cite{yu2018parallel} & dist. SGD & non-convex & \begin{tabular}{@{}c@{}}
	\textbf{\textit{non-IID}} \\ IID
\end{tabular} & \begin{tabular}{@{}c@{}}
synchronous \\   \textbf{\textit{asynchronous}}
\end{tabular} & full & - \\
\hline
\cite{wang2018adaptive} & dist. SGD & convex & \textit{\textbf{non-IID}} & synchronous & full & bounded gradients \\
\hline
\cite{yu19} & dist. momentum SGD & non-convex & \textbf{\textit{non-IID}} & synchronous & full & - \\
\hline
\cite{xie2019asynchronous} & FedAvg & \begin{tabular}{@{}c@{}}
convex \\   some non-convex
\end{tabular} & \textit{\textbf{non-IID}} & \textbf{\textit{asynchronous}} & full & -\\
\hline
\cite{li2019convergence} & FedAvg & convex & \textbf{\textit{non-IID}} & synchronous & \textbf{\textit{partial}} & bounded gradients
\\ \hline
\rowcolor{LightCyan}
\emph{This work} & FedAvg & convex & \textbf{\textit{non-IID}} & \textbf{\textit{asynchronous}} & \textbf{\textit{partial}} & model drift \\
\hline
\end{tabular}
\label{tab:ref}
\end{center}
\end{table*}

\noindent Returning to~\eqref{eq:globalProb}, in the absence of communication or computational constraints, the centralized gradient descent step takes the form:
\begin{equation}\label{eq:centStep}
	w_i = w_{i-1} - \mu \frac{1}{K}\sum_{k=1}^K\grad{w}P_k(w_{i-1}),
\end{equation}
where $i$ is the iteration index. We can distribute the update by splitting the gradient descent step among the $K$ agents. After introducing local parameters $\w_{k,i}$ at each agent, recursion \eqref{eq:centStep} becomes:
\begin{align}\label{eq:distStep1}
	w_{k,i} &= w_{i-1} - \mu\grad{w} P_k(w_{i-1}), \\ 
	\label{eq:distStep2}
	w_i &= \frac{1}{K}\sum_{k=1}^K w_{k,i-1}.
\end{align}
The drawback of this formulation is that evaluation of \( \grad{w} P_k(w_{i-1}) \) at each agent \( k \) and for every iteration \( i \) requires a total of \( N_k \) evaluations of the gradients \( \nabla Q_k(w_{i-1};\bm{x}_{k, n}) \). A popular approach for reducing the per-iteration computational cost is the utilization of stochastic gradient approximations. We will construct the gradient approximation as the average of \( E_k \) individual mini-batch approximations, each of size \( B_k \):
\begin{equation}\label{eq:gradient_approx}
	\widehat{\grad{w}P}_k(\w_{i-1}) \triangleq \frac{1}{E_k}\sum_{e=0}^{E_k-1}\frac{1}{B_k}\sum_{b\in \mathcal{\bm{B}}_{k, e}} \grad{w}Q_k(\w_{i-1};\bm{x}_{k,b}),
\end{equation}
where \( \mathcal{\bm{B}}_{k, e} \) denotes the \( e \)-th mini-batch set randomly sampled at time \( i \) by agent \( k \). We sample the indices in \( \mathcal{\bm{B}}_{k, e} \) from the set of integers \( \left\{ 1, \ldots, N_k  \right\} \) \emph{without replacement}. In the above, we are using the boldface notation $\w_{i-1}$ to reflect the random nature of the weight iterates. Note that this construction allows for significant heterogeneity in the agents' computational capabilities. In particular, by choosing \( E_k \) and \( B_k \) appropriately, each agent \( k \) is able to contribute to varying degrees, by performing a different number of gradient calculations. Then, the resulting stochastic gradient steps in \eqref{eq:distStep1}-\eqref{eq:distStep2} become:
\begin{align}\label{eq:distSGD}
	\w_{k,i} &= \w_{k,i-1} - \frac{\mu}{E_k B_k}\sum_{e=0}^{E_k-1}\sum_{b\in \mathcal{\bm{B}}_{k,  e}} \grad{w}Q_k(\w_{i-1};\bm{x}_{k,b}), \\
	\w_i &= \frac{1}{K}\sum_{k=1}^K \w_{k,i}\label{eq:distSGD2}.
\end{align}
An equivalent way of writing~\eqref{eq:distSGD}--\eqref{eq:distSGD2} is by introducing an inner iteration over \( e = 0, \ldots, E_{k}-1 \), initialized at \( \w_{k,-1} = \w_{i-1} \):
\begin{align}\label{eq:non_inc}
	\w_{k,e} &= \w_{k,e-1} - \frac{\mu}{E_k B_k} \sum_{b\in \mathcal{\bm{B}}_{k, e}} \grad{w}Q_k(\w_{i-1};\bm{x}_{k,b}),
\end{align}
followed by:
\begin{equation}\label{eq:full_part_comb}
  \w_i = \frac{1}{K} \sum_{k=1}^K \w_{k,E_k}.
\end{equation}
Examination of~\eqref{eq:non_inc} reveals that, while \( \w_{k, e} \) evolves with \( e \), all gradients \( \grad{w}Q_k(\w_{i-1};\bm{x}_{k,b}) \) are evaluated at \(\w_{k,-1} = \w_{i-1} \), which is the starting point of the inner loop. We can instead appeal to an incremental arguement~\cite{Bertsekas96anew} and replace~\eqref{eq:non_inc} by:
\begin{align}\label{eq:inc}
	\w_{k,e} &= \w_{k,e-1} - \frac{\mu}{E_k B_k} \sum_{b\in \mathcal{\bm{B}}_{k, e}} \grad{w}Q_k(\w_{k, e-1};\bm{x}_{k,b}),
\end{align}
where now all gradients \( \grad{w}Q_k(\w_{k, e-1};\bm{x}_b) \) are evaluated at the latest estimate \( \w_{k, e-1} \) at agent \( k \). Since full participation of all \( K \) agents at every time instant \( i \) is generally infeasible in a federated learning scenario~\cite{mcmahan16}, we allow for a participation of only \( L \) agents at every iteration, and sample the set of indices of participating agents \( \mathcal{L}_i \) from \(\{ 1, \ldots, L\} \) without replacement, transforming the combination step~\eqref{eq:full_part_comb} to:
\begin{equation}\label{eq:partial_part_comb}
  \w_i = \frac{1}{L} \sum_{\ell \in \mathcal{L}_i} \w_{\ell,E_\ell}.
\end{equation}
With the local incremental update step~\eqref{eq:inc} and partial-participation combination step~\eqref{eq:partial_part_comb} we arrive at Algorithm~\ref{alg:FL}.

This algorithm bears similarity to the original FedAvg~\cite{mcmahan16} algorithm, but differs in the fact that we allow for varying local epoch sizes $E_k$ and the normalization of gradient directions by epoch size. The advantage of this construction is in its applicability to settings where agents have varying capabilities so that some agents are able to run multiple epochs while others are not. Instead of forcing capable agents to only run a single epoch, not taking advantage of their computational resources, or forcing slow agents to perform multiple epochs, risking a straggler effect, this model allows every agent to contribute precisely as much as they are able to. The normalization of gradients by the number of local epochs \( E_k \), as our analysis will show, is necessary to ensure an unbiased solution despite asymmetric agent contribution, essentially by ensuring that~\eqref{eq:gradient_approx} is an unbiased estimate of the gradient of \( P_k(\cdot) \). In the absence of step-size normalization, agents with more participation would be able to bias the limiting point of the algorithm towards their own local minimizers.

\vspace*{0.2cm}
\begin{algorithm}
\begin{algorithmic}
\caption{(Dynamic Federated Averaging )}\label{alg:FL}
\STATE{
\textbf{initialize} $w_{0}$\;}
\FOR{each iteration $i=1,2,\cdots$}\STATE{
Select set of participating agents $\Li$ by sampling \( L \) times from \( \{ 1, \ldots, K \} \) without replacement.\\
\FOR{each agent $k \in \mathcal{L}_i$} \STATE {
\textbf{initialize} $\w_{k,-1} = \w_{i-1}$ \\
\FOR{each epoch $e=1,2,\cdots E_{k}$}\STATE{
Find indices of the mini-batch sample \( \mathcal{B}_{k,e} \) by sampling \( B_{k} \) times from \( \{ 1, \ldots, N_{k} \} \) without replacement.\\
$\bm{g}  =\frac{1}{B_k} \sum\limits_{b\in \mathcal{B}_{k, e}} \grad{w}Q(\w_{k,e-1};\bm{x}_{k,b})$ \\
$\w_{{k},e} = \w_{{k},e-1} - \mu\frac{1}{E_{k}}\bm{g}$ \\
}\ENDFOR
} \ENDFOR
\\ $\w_i = \frac{1}{L}\sum\limits_{k \in \Li} \w_{k,E_{k}}$
}\ENDFOR

\end{algorithmic}
\end{algorithm}

\section{Convergence Analysis}
\subsection{Modeling Conditions}
\noindent To facilitate the performance analysis of the dynamic federated averaging algorithm in a non-stationary environment, we assume convexity and smoothness of gradients.
\begin{assumption}\label{assum:convex}
	The functions $P_k(\cdot)$ are $\nu-$strongly convex, and $Q_k(\cdot; \boldsymbol{x}_{k, n})$ are convex:
	\begin{align}
		P_k(w_2) &\geq P_k(w_1) + \grad{w}P_k(w_1)(w_2-w_1) + \frac{\nu}{2}\Vert w_2-w_1\Vert^2,	 \\
		Q_k(w_2) &\geq Q_k(w_1) + \grad{w}Q_k(w_1)(w_2-w_1).
	\end{align}
	They also have $\delta-$Lipschitz gradients:
	\begin{align}
		\Vert \grad{w}P_k(w_2) - \grad{w} P_k(w_1)\Vert &\leq \delta \Vert w_2-w_1\Vert, \\
		\Vert \grad{w}Q_k(w_2; \boldsymbol{x})) - \grad{w} Q_k(w_1; \boldsymbol{x}))\Vert &\leq \delta \Vert w_2-w_1\Vert.
	\end{align}	\qed
\end{assumption}
\noindent We also impose an assumption on the drift of the minimizer \( \w_i^o \), namely that it follows a random walk model.
\begin{assumption}\label{assum:drift}
We assume that the true model \( \w^o_i \) follows a random walk:
\begin{equation}\label{eq:driftModel}
	\w^o_i = \w^o_{i-1}  + \bm{q}_i,
\end{equation}
where $\bm{q}_i$ denotes some zero mean random variable independent of $\w^o_j$ for any $j < i$ and with bounded variance, i.e., $\mathbb{E} \Vert \bm{q}_i \Vert^2 = \sigma_q^2$.	\qed
\end{assumption}
\noindent As it will turn out, the tracking performance of Algorithm~\ref{alg:FL} will be largely determined by the drift parameter \( \sigma_q^2 \) on the \emph{global} model \( \w_i^o \). Nevertheless, we need to additionally assume that the individual minimizers \( \w_{k, i}^o \triangleq \argmin\limits_{w \in \mathbb{R}^M} P_k(w) \) do not drift too far.
\begin{assumption}\label{assum:localdrift}
For all \( i \), the distance of each local model \( \w_{k, i}^o \) to the global model \( \w_{i}^o \) is bounded, i.e.:
\begin{equation}
  \mathbb{E} {\|\w_{k, i}^o - \w_{i}^o \|}^2 \le \sigma_{w, k}^2.
\end{equation}
\qed
\end{assumption}

\begin{figure*}[htbp]
	\begin{subfigure}{.25\textwidth}
  \centering
  \includegraphics[width=\linewidth]{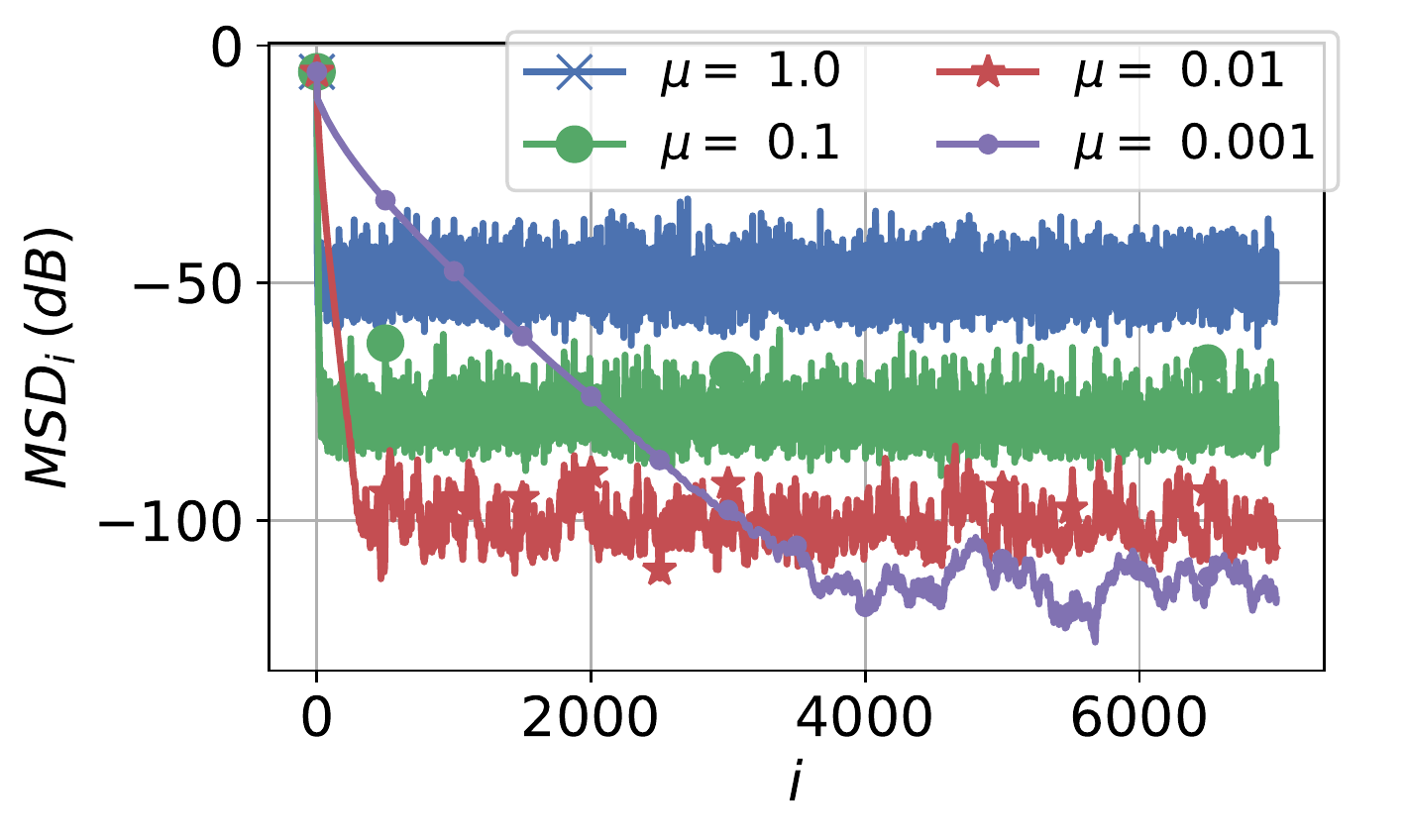}
  \caption{\textit{Stationary case:} varying $\mu$ }
  \label{fig:mu}
\end{subfigure}%
\begin{subfigure}{.25\textwidth}
  \centering
  \includegraphics[width=\linewidth]{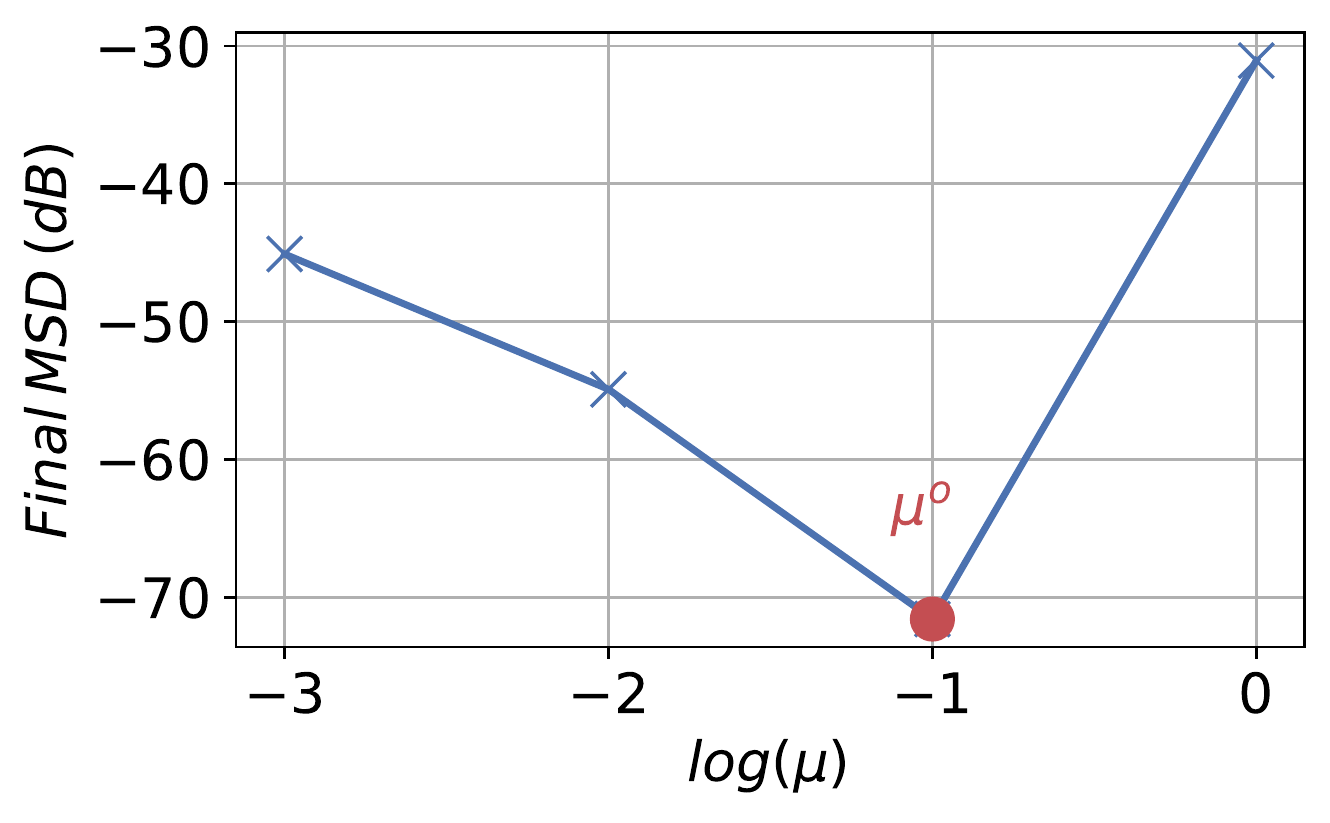}
  \caption{\textit{Non-stationary case:} varying $\mu$ }
  \label{fig:mu-non}
\end{subfigure}
\begin{subfigure}{.25\textwidth}
  \centering
  \includegraphics[width=\linewidth]{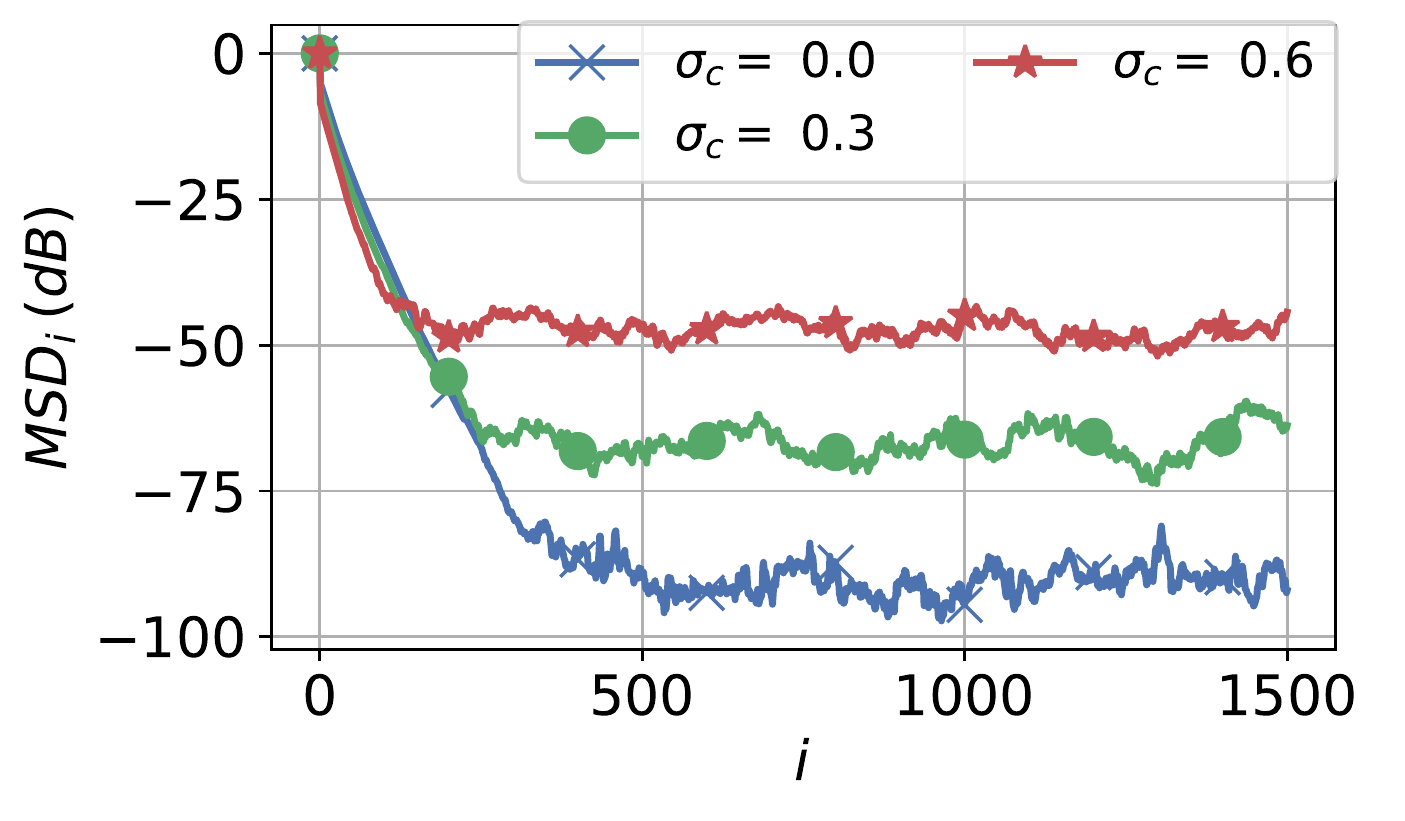}
  \caption{Varying $\sigma_c^2$}
  \label{fig:sigc}
\end{subfigure}%
\begin{subfigure}{.25\textwidth}
  \centering
  \includegraphics[width=\linewidth]{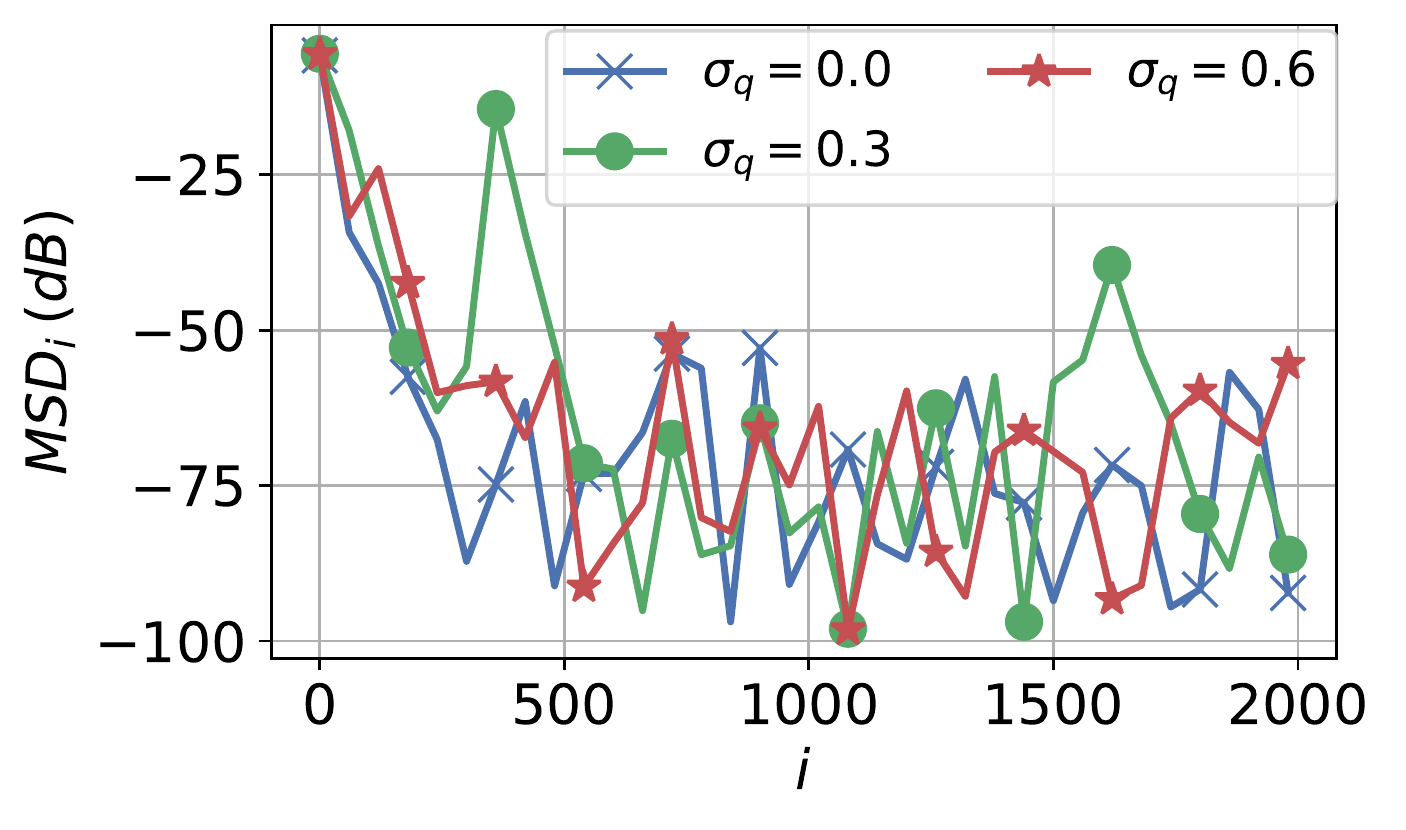}
  \caption{Varying $\sigma_q^2$}
  \label{fig:sigq}
\end{subfigure} %
\caption{Plots of the MSD across time in dB. }
\label{fig:fig}
\end{figure*}

\subsection{Error Recursion}
\noindent Iterating the local update steps~\eqref{eq:inc} and combining via~\eqref{eq:partial_part_comb}, we find:
\begin{align}\label{eq:fedrec}
	\w_i =&\: \w_{i-1} - \mu \frac{1}{L}\sum_{\ell \in \mathcal{L}_i} \frac{1}{E_{\ell} B_{\ell}}\sum_{e=0}^{E_{\ell}-1}\sum_{b\in \mathcal{\bm{B}}_{\ell,e}} \grad{w}Q_{\ell}(\w_{\ell,e-1};\bm{x}_{\ell,b}) \notag\\
	=&\: \w_{i-1} - \mu \frac{1}{K}\sum_{k=1}^K\grad{w}P_k(\w_{i-1}) - \mu \boldsymbol{s}_i - \mu \boldsymbol{d}_i,
\end{align}
where we introduced:
\begin{align}
  \si \triangleq&\: \frac{1}{L}\sum_{\ell \in \Li}\widehat{\grad{w}P}_{\ell}(\w_{i-1}) - \frac{1}{K}\sum_{k=1}^K \grad{w}P_k(\w_{i-1}), \label{eq:gradNoise}\\
  \boldsymbol{d}_i \triangleq&\: \frac{1}{L}\sum_{\ell \in \Li} \frac{1}{E_{\ell}B_{\ell}}\sum_{e=0}^{E_{\ell}-1}\sum_{b\in \mathcal{\bm{B}}_{\ell,  e}} \Big[ \grad{w}Q_{\ell}(\w_{i-1};\bm{x}_{\ell,b}) \notag \\
	& \ \ \ \ \ \ \ \ \ \ \ \ \ \ \ \ \ \ \ \ \ \ \ \ \ \ \ \ \ \ \ \ \   - \grad{w}Q_{\ell}(\w_{\ell,e};\bm{x}_{\ell,b}) \Big].
\end{align}
The term \( \boldsymbol{s}_i \) results from the stochastic approximation of the true gradient of~\eqref{eq:globalProb} by only utilizing a subset of agents and a subset of data at every iteration. The term \( \boldsymbol{d}_i \) results from the incremental implementation of~\eqref{eq:inc}. The gradient noise term \( \boldsymbol{s}_i \) will turn out to be the dominant factor in the performance of the algorithm, but can be bounded as follows.
\begin{lemma}\label{lem:gradNoise}
	The gradient noise defined in \eqref{eq:gradNoise} satisfies:
	\begin{align}
		\mathbb{E}\: \left\{ \bm{s}_i | {\w_{i-1}} \right\} &= 0, \label{eq:mean_zero}\\
		\mathbb{E} \left\{ \Vert \si\Vert^2 | \w_{i-1} \right\} & \leq \beta_s^2  \mathbb{E} \Vert \w^o_{i-1}-\w_{i-1} \Vert^2 + \sigma_s^2 + \epsilon^2, \label{eq:variance_bound}
	\end{align}
	where we defined:
	\begin{align}
		\beta_s^2 &\eqdef  \frac{1}{KL} \sum_{k=1}^K \Big( 6\tau_{s,k} + 2\tau_{\epsilon} \Big)	\delta^2, \\
		\sigma_s^2 &\eqdef \frac{3}{KL} \sum_{k=1}^K  \tau_{s, k} \mathbb{E}\Vert \grad{w} Q_k(\w_{i-1}^o;\bm{x}_{k}) \notag \\
		& \ \  \ \ \ \ \ \ \ \ \ \ \ \ \ \ \ \ \ \  \ \ \ \	-\grad{w}P_k(\w_{i-1}^o)\Vert^2,  \\
    \epsilon^2 &\eqdef \frac{2}{KL} \sum_{k=1}^K \tau_{\epsilon}^2	 \mathbb{E} \Vert \grad{w}P_k(\w_{i-1}^o)\Vert^2, \\
		\tau_{s, k} &\eqdef \frac{N_k-B_k}{(N_k-1)B_kE_k }, \quad	\tau_{\epsilon} \eqdef \frac{K-L}{K-1}.
	\end{align}
\end{lemma}
\begin{proof}
	Omitted due to space limitations.
\end{proof}
\noindent From~\eqref{eq:variance_bound}, we observe that the bound on the gradient noise variance consists of two absolute components, \( \sigma_s^2 \) and \( \epsilon^2 \). The term \( \sigma_s^2 \) corresponds to an average of the \emph{data variability} \( \mathbb{E}\Vert \grad{w} Q_k(\w_{i-1}^o;\bm{x}_{k}) -\grad{w}P_k(\w_{i-1}^o)\Vert^2 \) at each agent, weighted by the factor \( \tau_{s, k} \). Agents with high data variability will incur a higher variance by employing a mini-batch approximation, instead of a full gradient update, but can mitigate this effect by increasing the mini-batch size \( B_k \), and hence reducing \( \tau_{s, k} \). In the limit case where agent \( k \) is performing a full gradient update, we have \( B_k = N_k \), and hence \( \tau_{s, k} = 0 \) and no noise contribution from agent \( k \). The second absolute noise term \( \epsilon^2 \) stems from \emph{model variability} among different agents. In particular, \( \mathbb{E} \Vert \grad{w}P_k(\w_{i-1}^o)\Vert^2 \) measures the suboptimality of the average model \( \w_{i-1}^o \) for the local cost \( P_k(\cdot) \) at agent \( k \), and in light of the Lipschitz gradient condition is proprtional to \(\mathbb{E} \|\w_{i-1}^o - \w_{k, i-1}^o\|^2\); it stems from the incremental step introduced in~\eqref{eq:inc}. In this case, the model variability term is multiplied by a participation factor \( \tau_{\epsilon} \), which quantifies the fraction of agents participating in the federated update at every iteration, and vanishes whenever all agents participate. 

\begin{theorem}\label{thrm}
	Consider the iterates \( \w_i \) generated by the dynamic federated averaging algorithm. For sufficiently small step-size \( \mu \),
it holds that $\mathbb{E}\Vert \w_i^o - \w_i \Vert^2$ converges exponentially fast:
\begin{align}\label{eq:MSDBound}
	\mathbb{E}\Vert \w_i^o - \w_i \Vert^2 \leq O(\gamma^i) + O(\mu)\left( \sigma_s^2 + \epsilon^2\right) + O(\mu^{-1})\sigma_q^2,
\end{align}
where \(\gamma < 1-\nu\mu + O(\mu^2) \in [0,1)\).

\end{theorem}
\begin{proof}
	Omitted due to space limitations.
\end{proof}
\noindent We observe that, in addition to the model and data variability terms discussed above, the performance of the algorithm is further determined by the drift parameters of the random-walk model \( \sigma_q^2 \).
A reduction in the step-size results in slower decay of the transient term \( O(\gamma^i) \) and increased tracking loss \( O(\mu^{-1}) \sigma_q^2 \), while reducing the effect of the gradient noise proportional to \( O(\mu) \).

\section{Experimental Analysis}

\subsection{Experimental Setup}
We test the dynamic federated averaging algorithm on the logistic risk function with $\ell_2-$norm regularization \eqref{eq:costfct} using synthetic data. The local risk is given by:
\begin{equation}\label{eq:costfct}
	P_k(w) = \frac{1}{N_k}\sum_{n=0}^{N_k-1} \text{ln}(1+e^{-\gamma h^T w}) + \rho \Vert w \Vert^2.
\end{equation}
The data is generated as follows: We first generate a random $\w^{\star}_0$ and apply the random walk model \eqref{eq:driftModel} with $\bm{q}_i \sim \mathcal{N}(0,\sigma_q^2)$. Then, we model the change in the true parameters $\w^{\star}_i$ across agents by adding randomly sampled constants $\bm{c}_k\sim \mathcal{N}(0,\sigma_c^2)$, i.e., $\w^{\star}_{k,i} = \w^{\star}_i + \bm{c}_k$. Then, for each agent $k$, $N_k $ random $2-$dimensional features are sampled from a Gaussian distribution with a randomly generated variance, for each time $i$.  We generate random feature vectors $\bm{h}_{k,i}$ and assign them labels $\bm{\gamma}_{k,i} =\text{sign}( \bm{h}_{k,i}^T \w^{\star}_{k,i})$.  We assume we have $K = 20$ agents, with $L = 7$ active agents at each time. We set the batch sizes $B_k$ and epoch sizes $E_k$ to different values in the range $[10,20]$ and $[1,10]$, respectively. 

We examine the effect of different hyperparameters on the behaviour of the algorithm. We validate the theoretical results by changing the step size; we also look at the effect of varying the variances $\sigma_q^2$ and $\sigma_c^2$. We plot the averaged mean-square-deviation (MSD) curves in log domain after performing 50 passes over the data (Figure \ref{fig:fig}). For the first experiment, we set $\sigma_c^2=0.1$, and we vary the step size by factor of 10 in the stationary ($\sigma_q^2=0$) and non-stationary case ($\sigma_q^2=0.01$). We observe in Figure \ref{fig:mu} that as $\mu$ increases, the MSD increases and the convergence becomes faster; while, in Figure \ref{fig:mu-non} we plot the final MSD value reached for each $\mu$, and we observe that there is an optimal step size $\mu^o = 0.1$ that achieves the best MSD. For the second experiment, we vary $\sigma_c^2$ while fixing $\sigma_q^2=0$ and $\mu = 0.01$. The plots in Figure \ref{fig:sigc} indicate that the MSD increases with $\sigma_c^2$, but the convergence rate is not effected, which is expected. Finally, in the third experiment we fix $\sigma_c^2=0.1$ and vary $\sigma_q^2$. A similar trend is observed (Figure \ref{fig:sigq}); $\sigma_q^2$ only affects the MSD and not the convergence.

\section{Conclusion}
The work presented in this paper consists of developing a convergence analysis for a modified version of the FedAvg algorithm under three major challenges: non-IID data, asynchronous operation under drift, and partial agent participation. We were able to guarantee the convergence of the algorithm and identified three major components that affect its convergence: step size, agent heterogeneity, and drift variance. We were able to illustrate the theoretical results with  a series of experiments. 
\newpage

\bibliographystyle{IEEEtran}
{\balance{\bibliography{SPAWC-refs}}}

\end{document}